\documentclass[11pt,a4paper]{article}
\usepackage[hyperref]{naaclhlt2018}
\usepackage{times}
\usepackage{latexsym}

\aclfinalcopy
\def\aclpaperid{341} 
\usepackage{url}

\usepackage{graphicx} 
\usepackage{subfigure}

\usepackage{textcomp}

\usepackage{multirow}
\usepackage{bm,color}
\usepackage{mathdots}
\usepackage{amsmath, amssymb,amsthm}

\usepackage{graphicx}
\usepackage{caption}
\usepackage{footnote}

\makesavenoteenv{table}
\makesavenoteenv{tabular}

\newcommand{\R}{\mathbb{R}}

\newcommand{\cv}{\boldsymbol{c}}

\newcommand{\uv}{\boldsymbol{u}}
\newcommand{\vv}{\boldsymbol{v}}

\newcommand{\Um}{\boldsymbol{U}}
\newcommand{\Vm}{\boldsymbol{V}}

\newcommand{\id}{\boldsymbol{\iota}}

\newcommand{\vocab}{\mathcal{V}}
\newcommand{\corp}{\mathcal{C}}

\DeclareMathOperator*{\argmin}{arg\,min}

\newcommand\blfootnote[1]{%
  \begingroup
  \renewcommand\thefootnote{}\footnote{#1}%
  \addtocounter{footnote}{-1}%
  \endgroup
}

\title{Unsupervised Learning of Sentence Embeddings\\ using Compositional n-Gram Features}

\author{Matteo Pagliardini* \\
  Iprova SA, Switzerland \\
  {\small\tt mpagliardini@iprova.com } \\\And
  Prakhar Gupta* \\
  EPFL, Switzerland \\
  {\small\tt prakhar.gupta@epfl.ch}  \\\And
  Martin Jaggi \\
  EPFL, Switzerland \\
  {\small\tt martin.jaggi@epfl.ch} \\}

\date{}

\begin{document}
\maketitle
\begin{abstract}
The recent tremendous success of unsupervised word embeddings in a multitude of applications raises the obvious question if similar methods could be derived to improve embeddings (i.e. semantic representations) of word sequences as well.
We present a simple but efficient unsupervised objective to train distributed representations of sentences.
Our method outperforms the state-of-the-art unsupervised models on most benchmark tasks, highlighting the robustness of the produced general-purpose sentence embeddings.
\end{abstract}

\section{Introduction}
\blfootnote{* indicates equal contribution}
Improving unsupervised learning is of key importance for advancing machine learning methods, as to unlock access to almost unlimited amounts of data to be used as training resources.
The majority of recent success stories of deep learning does not fall into this category but instead relied on supervised training (in particular in the vision domain).
A very notable exception comes from the text and natural language processing domain, in the form of semantic word embeddings trained unsupervised \cite{Mikolov:2013uz,mikolov2013efficient,pennington2014glove}.
Within only a few years from their invention, such word representations -- which are based on a simple matrix factorization model as we formalize below -- are now routinely trained on very large amounts of raw text data, and have become ubiquitous building blocks of a majority of current state-of-the-art NLP applications.

\vspace{0.25em}
While very useful semantic representations are available for words, it remains challenging to produce and learn such semantic embeddings for longer pieces of text, such as sentences, paragraphs or entire documents. Even more so, it remains a key goal to learn such general-purpose representations in an unsupervised way.

Currently, two contrary research trends have emerged in text representation learning: On one hand, a strong trend in deep-learning for NLP leads towards increasingly powerful and complex models, such as recurrent neural networks (RNNs), LSTMs, attention models and even Neural Turing Machine architectures.  %
While extremely strong in expressiveness, the increased model complexity makes such models much slower to train on larger datasets%
. On the other end of the spectrum, simpler ``shallow'' models such as matrix factorizations (or bilinear models) can benefit from training on much larger sets of data, which can be a key advantage, especially in the unsupervised setting.

\vspace{0.25em}
Surprisingly, for constructing sentence embeddings, naively using averaged word vectors was shown to outperform LSTMs (see \citet{wieting2016} for plain averaging, and \citet{arora2017} for weighted averaging). %
This example shows potential in exploiting the trade-off between model complexity and ability to process huge amounts of text using scalable algorithms, towards the simpler side.
In view of this trade-off, our work here further advances unsupervised learning of sentence embeddings.
Our proposed model can be seen as an extension of the C-BOW \cite{Mikolov:2013uz,mikolov2013efficient} training objective to train sentence instead of word embeddings. We demonstrate that the empirical performance of our resulting general-purpose sentence embeddings very significantly exceeds the state of the art, while keeping the model simplicity as well as training and inference complexity exactly as low as in averaging methods \cite{wieting2016,arora2017}, thereby also putting the work by \cite{arora2017} in perspective.

\vspace{0.35em}
\textbf{Contributions.} The main contributions in this work can be summarized as follows:

\begin{itemize}%

 \item \textbf{Model.} We propose \textbf{Sent2Vec}\footnote{
 All our code and pre-trained models are publicly available for download at
 \url{http://github.com/epfml/sent2vec}}, a simple unsupervised model allowing to compose sentence embeddings using word vectors along with n-gram embeddings, simultaneously training composition and the embedding vectors themselves.

 \item \textbf{Efficiency \& Scalability.}
 The computational complexity of our embeddings is only $\mathcal{O}(1)$ vector operations per word processed, both during training and inference of the sentence embeddings. This strongly contrasts all neural network based approaches, and allows our model to learn from extremely large datasets, in a streaming fashion, which is a crucial advantage in the unsupervised setting. Fast inference is a key benefit in downstream tasks and industry applications.
 \item \textbf{Performance.} Our method shows significant performance improvements compared to the current state-of-the-art unsupervised and even semi-supervised models. The resulting general-purpose embeddings show strong robustness when transferred to a wide range of prediction benchmarks.

\end{itemize}

\section{Model}
\label{sec:model}

Our model is inspired by simple matrix factor models (bilinear models) such as recently very successfully used in unsupervised learning of word embeddings \cite{Mikolov:2013uz,mikolov2013efficient,pennington2014glove,Bojanowski:2017un} as well as supervised %
of sentence classification \cite{Joulin:2017uo}.
More precisely, these models can all be formalized as an optimization problem of the form%
\begin{equation}\label{eq:fact}
\min_{\Um,\Vm} ~~ \sum_{S\in\corp} f_S(\Um\Vm\id_S)
\end{equation}
for two parameter matrices $\Um\in\R^{k \times h}$ and $\Vm\in\R^{h\times |\vocab|}$, where $\vocab$ denotes the vocabulary. Here, the columns of the matrix $\Vm$ represent the learnt source word vectors whereas those of $\Um$ represent
the target word vectors.
For a given sentence~$S$, which can be of arbitrary length, the indicator vector $\id_S \in \{0,1\}^{|\vocab|}$ is a binary vector encoding $S$ (bag of words encoding).

Fixed-length context windows $S$ running over the corpus are used in word embedding methods as in C-BOW \cite{Mikolov:2013uz,mikolov2013efficient} and GloVe \cite{pennington2014glove}. Here we have $k=|\vocab|$ and each cost function $f_S: \R^k\rightarrow\R$ only depends on a single row of its input, describing the observed target word for the given fixed-length context~$S$.
In contrast, for sentence embeddings which are the focus of our paper here, $S$ will be entire sentences or documents (therefore variable length). This property is shared with the supervised FastText classifier \cite{Joulin:2017uo}, which however uses soft-max with $k\ll|\vocab|$ being the number of class labels.

\subsection{Proposed Unsupervised Model}
We propose a new unsupervised model, \textbf{Sent2Vec}, for learning universal sentence embeddings.
Conceptually, the model can be interpreted as a natural extension of the word-contexts from
C-BOW \cite{Mikolov:2013uz,mikolov2013efficient}  to a larger sentence context,
with the sentence words being specifically optimized towards additive combination over the sentence, by means of the
unsupervised objective function.

Formally, we learn a source (or context) embedding $\vv_w$ and target embedding $\uv_w$ for each word~$w$ in the vocabulary, with embedding dimension $h$ and $k=|\vocab|$ as in \eqref{eq:fact}. The sentence embedding is defined as the average of the source word embeddings of its constituent words, as in \eqref{eq:sentEmb}.
We augment this model furthermore by also learning source embeddings for not only unigrams but also n-grams present in each sentence, and averaging the n-gram embeddings along with the words, i.e., the sentence embedding $\vv_S$ for $S$ is modeled as%
\begin{align}\label{eq:sentEmb}
\vv_{S} := \tfrac{1}{|R(S)|}  \Vm \id_{R(S)} = \tfrac{1}{|R(S)|} \sum_{w \in R(S)} \vv_w
\end{align}
where $R(S)$ is the list of n-grams (including unigrams) present in sentence $S$.
In order to predict a missing word from the context, our objective models the softmax output approximated by negative sampling following \cite{Mikolov:2013uz}.
For the large number of output classes $|\vocab|$ to be predicted, negative sampling is known to significantly improve training efficiency, see also \cite{Goldberg:2014uy}.
Given the binary logistic loss function $\ell: x \mapsto \log{(1 + e^{-x})}$ coupled with negative sampling, our unsupervised training objective is formulated as follows:
\begin{align*}
\min_{\Um,\Vm} \ \
\sum_{S \in \corp} & \sum_{w_t \in S}
\bigg( \ell\big(\uv_{w_t}^\top \vv_{S\setminus\{w_t\}}\big) \\
& + \!\!\! \sum_{w' \in N_{w_t}} \!\!\! \ell\big(\!-\uv_{w'}^\top \vv_{S\setminus\{w_t\}}\big)\bigg) \notag
\end{align*}
where $S$ corresponds to the current sentence and $N_{w_t}$ is the set of words sampled negatively for the word $w_t \in S$. The negatives are sampled\footnote{To efficiently sample negatives, a pre-processing table is constructed, containing the words corresponding to the square root of their corpora frequency.
Then, the negatives $N_{w_t}$ are sampled uniformly at random from the negatives table except the target $w_t$ itself, following \cite{Joulin:2017uo,Bojanowski:2017un}.} following a multinomial distribution where each word~$w$ is associated with a probability $q_n(w) := \sqrt{f_w} \,\big/\, \big(\sum_{w_i \in \vocab} \sqrt{f_{w_i}}\big)$,
where $f_w$ is the normalized frequency of $w$ in the corpus.

To select the possible target unigrams (positives), we use subsampling as in \cite{Joulin:2017uo,Bojanowski:2017un},
each word $w$ being discarded with probability $1-q_p(w)$ where $q_p(w) :=\min \big\{1, \sqrt{ t / f_w } +  t / f_w  \big\}$.
Where~$t$ is the subsampling hyper-parameter.
Subsampling prevents very frequent words of having too much influence in the learning as they would introduce strong biases in the prediction task.
With positives subsampling and respecting the negative sampling distribution, the precise training objective function becomes
\begin{align}\label{eq:trainObj}
\min_{\Um,\Vm} \ \
\sum_{S \in \corp} & \sum_{w_t \in S}
\bigg( q_p(w_t) \ell\big(\uv_{w_t}^\top \vv_{S\setminus\{w_t\}}\big) \\
&  + |N_{w_t}| \!\sum_{w' \in \vocab} q_n(w') \ell\big(\!-\uv_{w'}^\top \vv_{S\setminus\{w_t\}}\big)\bigg) \notag
\end{align}

\subsection{Computational Efficiency}
In contrast to more complex neural network based models, one of the core advantages of the proposed technique is the low computational cost for both inference and training.
Given a sentence $S$ and a trained model, computing the sentence representation $\vv_{S}$ only requires $|S|\cdot h$ floating point operations (or $|R(S)|\cdot h$ to be precise for the n-gram case, see \eqref{eq:sentEmb}), where $h$ is the embedding dimension.
The same holds for the cost of training with SGD on the objective \eqref{eq:trainObj}, per sentence seen in the training corpus.
Due to the simplicity of the model, parallel training is straight-forward using parallelized or distributed SGD.

Also, in order to store higher-order n-grams efficiently, we use the standard hashing trick, see e.g. \cite{weinberger2009feature}, with the same hashing function as used in FastText \cite{Joulin:2017uo,Bojanowski:2017un}.

\subsection{Comparison to C-BOW}
\label{sec:C-BOW}

C-BOW \cite{Mikolov:2013uz,mikolov2013efficient} aims to predict a chosen target word given its fixed-size context window,
the context being defined by the average of the vectors associated with the words at a distance less than the window size hyper-parameter $ws$.
If our system, when restricted to unigram features, can be seen as an extension of C-BOW where the context window includes the entire sentence,
in practice there are few important differences as C-BOW uses important tricks to facilitate the learning of word embeddings.
C-BOW first uses frequent word subsampling on the sentences, deciding to discard each token $w$ with probability $q_p(w)$ or alike (small variations exist across implementations). %
Subsampling prevents the generation of n-grams features, and deprives the sentence of
an important part of its syntactical features. It also shortens the distance between subsampled words, implicitly increasing the
span of the context window. A second trick consists of using dynamic context windows: for each subsampled word $w$, the size of its associated
context window is sampled uniformly between 1 and $ws$.
Using dynamic context windows is equivalent to weighing by the distance from the
focus word $w$ divided by the window size \cite{levy2015improving}. This makes the prediction task local, and go against our
objective of creating sentence embeddings as we want to learn how to compose all n-gram features present in a sentence.
In the results section, we report a significant improvement of our method over C-BOW.

\subsection{Model Training}
Three different datasets have been used to train our models: the Toronto book corpus\footnote{\url{http://www.cs.toronto.edu/~mbweb/}}, Wikipedia sentences and tweets.
The Wikipedia and Toronto books sentences have been tokenized using the Stanford NLP library \cite{manning2014stanford}, while for tweets we used the NLTK tweets tokenizer \cite{bird2009natural}.
For training, we select a sentence randomly from the dataset and then proceed to select all the possible target
unigrams using subsampling. We update the weights using SGD with a linearly decaying learning rate.

Also, to prevent overfitting, for each sentence we use dropout on its list of n-grams $R(S)\setminus \{U(S)\}$,
where $U(S)$ is the set of all unigrams contained in sentence $S$.
After empirically trying multiple dropout schemes, we find that dropping $K$ n-grams ($n>1$) for each sentence is giving superior results compared to dropping each token with some fixed probability.
This dropout mechanism would negatively impact shorter sentences.
The regularization can be pushed further by applying
L1 regularization to the word vectors. Encouraging sparsity in the embedding vectors is particularly beneficial for high dimension~$h$.
The additional soft thresholding in every SGD step adds negligible computational cost. See also Appendix~\ref{l1Details}.
We train two models %
on each dataset, one with unigrams only and one with unigrams and bigrams.
All training parameters for the models are provided in Table \ref{table_train} in the supplementary material. %
Our C++ implementation builds upon the FastText library \cite{Joulin:2017uo,Bojanowski:2017un}.
We will make our code and pre-trained models available open-source.

\section{Related Work}
\label{sec:rel-work}

We discuss existing models which have been proposed to construct sentence embeddings. While there is a large body of works in this direction -- several among these using e.g. labelled datasets of paraphrase pairs to obtain sentence embeddings in a supervised manner \cite{wieting2016charagram,wieting2016,conneau2017supervised}
to learn sentence embeddings -- we here focus on unsupervised, task-independent models. %
While some methods require ordered raw text i.e., a coherent corpus where the next sentence is a logical continuation of the previous sentence,
others rely only on raw text i.e., an unordered collection of sentences. Finally, we also discuss alternative models built from structured data
sources.

\subsection{Unsupervised Models Independent of Sentence Ordering}

The \textbf{ParagraphVector DBOW} model \cite{le2014distributed} is a log-linear model which is trained to learn sentence as well as word
embeddings and then use a softmax distribution to predict words contained in the sentence given the sentence vector representation.
They also propose a different model \textbf{ParagraphVector DM} where they use n-grams of consecutive words along with the
sentence vector representation to predict the next word.

\cite{lev2015defense} also presented an early approach to obtain compositional embeddings from word vectors.
They use different compositional techniques including static averaging or Fisher vectors of a multivariate Gaussian to obtain sentence embeddings from word2vec models.

\citet{Hill:2016uu} propose a \textbf{Sequential (Denoising) Autoencoder, S(D)AE}. This model first introduces noise in the input data: %
Firstly each word is deleted with probability $p_0$, then for each non-overlapping bigram, words are swapped with probability $p_x$. The model then uses an LSTM-based architecture to retrieve the original sentence from
the corrupted version. The model can then be used to encode new sentences into vector representations. In the case of $p_0 = p_x = 0$,
the model simply becomes a Sequential Autoencoder.
\citet{Hill:2016uu} also propose a variant \textbf{(S(D)AE + embs.)} in which the words are represented by fixed pre-trained word vector embeddings.

\citet{arora2017} propose a model in which sentences are represented as a weighted average of fixed (pre-trained) word vectors, followed by post-processing step of subtracting the principal component.
Using the generative model of \cite{Arora:2016vw}, words are generated conditioned on a sentence ``discourse'' vector~$\cv_s$:
\begin{align*}
Pr[w \,|\, \cv_s] = \alpha f_w + (1-\alpha)\frac{\exp( \tilde{\cv}_s^\top \vv_w)}{Z_{\tilde{\cv}_s}} ,
\end{align*}
where $Z_{\tilde{\cv}_s} := \sum_{w \in \vocab} \exp(\tilde{\cv}_s^\top \vv_w)$ and $\tilde{\cv}_s := \beta \cv_0 + (1-\beta)\cv_s$ and $\alpha$, $\beta$ are scalars.
$\cv_0$ is the common discourse vector, representing a shared component among all discourses, mainly related to syntax. It allows
the model to better generate syntactical features. The $\alpha f_w$ term is here to enable the model to generate some frequent words even if their matching
with the discourse vector $\tilde{\cv}_s$ is low.

Therefore, this model tries to generate sentences as a mixture of three type of words:
words matching the sentence discourse vector $\cv_s$, syntactical words matching $\cv_0$, and words with high~$f_w$. \cite{arora2017} demonstrated that for
this model, the MLE of $\tilde{\cv}_s$ can be approximated by $\sum_{w \in S} \frac{a}{f_w + a} \vv_w$, where $a$ is a scalar. The sentence discourse vector
can hence be obtained by subtracting $\cv_0$ estimated by the first principal component of $\tilde{\cv}_s$'s on a set of sentences.
In other words, the sentence embeddings are obtained by a weighted average of the word vectors stripping away the syntax by subtracting the common discourse vector and down-weighting frequent tokens.
They generate sentence embeddings from diverse pre-trained word embeddings among
which are unsupervised word embeddings such as GloVe \cite{pennington2014glove} as well as supervised word embeddings such as paragram-SL999~(PSL) \cite{wieting2015paraphrase} trained on the
Paraphrase Database \cite{ganitkevitch2013ppdb}.

In a very different line of work, \textbf{C-PHRASE} \cite{pham2015jointly} relies on additional information from the syntactic parse tree of each sentence, which is incorporated into the C-BOW training objective.

\citet{Huang:2016uk} show that single layer CNNs can be modeled using a tensor decomposition approach. While building on an unsupervised objective, the employed dictionary learning step for obtaining phrase templates is task-specific (for each use-case), not resulting in general-purpose embeddings.

\subsection{Unsupervised Models Depending on Sentence Ordering}
The \textbf{SkipThought} model \cite{Kiros:2015uq} combines sentence level models with recurrent neural networks. Given a sentence
$S_i$ from an ordered corpus, the model is trained to predict $S_{i-1}$ and~$S_{i+1}$.

\textbf{FastSent} \cite{Hill:2016uu} is a sentence-level log-linear bag-of-words model. Like SkipThought, it uses adjacent sentences as the prediction target and is trained in an unsupervised fashion.
Using word sequences allows the model to improve over the earlier work of paragraph2vec \cite{le2014distributed}.
\cite{Hill:2016uu} augment FastSent further by training it to predict the constituent words of the sentence as well. This model is named \textbf{FastSent + AE} in our comparisons.

Compared to our approach, \textbf{Siamese C-BOW} \cite{kenter2016siamese} shares the idea of learning to average word embeddings over a sentence. However, it relies on a Siamese neural network architecture to predict surrounding sentences, contrasting our simpler unsupervised objective.

Note that on the character sequence level instead of word sequences, FastText \cite{Bojanowski:2017un} uses
the same conceptual model to obtain better word embeddings.
This is most similar to our proposed model, with two key differences:
Firstly, we predict from source word sequences to target words, as opposed to character sequences to target words,
and secondly, our model is averaging the source embeddings instead of summing them.

\subsection{Models requiring structured data}
\textbf{DictRep} \cite{Hill16TACL} is trained to map dictionary definitions of the words to the pre-trained word embeddings of
 these words. They use two different architectures, namely BOW and RNN (LSTM) with the choice of learning the input word embeddings or using them pre-trained.
A similar architecture is used by the \textbf{CaptionRep} variant, but here the task is the mapping of given image captions to a pre-trained vector representation of these images.

\section{Evaluation Tasks}

We use a standard set of supervised as well as unsupervised benchmark tasks from the literature to evaluate our trained models, following \cite{Hill:2016uu}. The breadth of tasks allows to fairly measure generalization to a wide area of different domains, testing the general-purpose quality (universality) of all competing sentence embeddings.
For downstream supervised evaluations, sentence embeddings are combined with logistic regression to predict target labels.
In the unsupervised evaluation for sentence similarity, correlation of the cosine similarity between two embeddings is compared to human annotators.

\textbf{Downstream Supervised Evaluation.}
Sentence embeddings are evaluated for various supervised classification tasks as follows.
We evaluate paraphrase identification (MSRP) \cite{dolan2004unsupervised}, classification of movie review sentiment~(MR) \cite{pang2005seeing}, product reviews (CR) \cite{hu2004mining}, subjectivity classification (SUBJ) \cite{pang2004sentimental}, opinion polarity
(MPQA) \cite{wiebe2005annotating} and question type classification (TREC) \cite{voorhees2002}.
To classify, we use the code provided by \cite{Kiros:2015uq} in the same manner as in \cite{Hill:2016uu}.
For the MSRP dataset, containing pairs of sentences $(S_1,S_2)$ with associated paraphrase label, we generate feature vectors by concatenating their
Sent2Vec representations $|\vv_{S_1} - \vv_{S_2}|$ with the component-wise product $\vv_{S_1}\odot \vv_{S_2}$.
The predefined training split is used to tune the L2 penalty parameter using cross-validation and the accuracy and F1 scores are computed on the test set.
For the remaining 5 datasets, Sent2Vec embeddings are inferred from input sentences and directly fed to a logistic regression classifier.
Accuracy scores are obtained using 10-fold cross-validation for the MR, CR, SUBJ and MPQA datasets. For those datasets nested cross-validation is used to tune the L2 penalty.
For the TREC dataset, as for the MRSP dataset, the L2 penalty is tuned on the predefined train split using 10-fold cross-validation, and the accuracy is computed on the test set.

\textbf{Unsupervised Similarity Evaluation.}
We perform unsupervised evaluation of the  learnt sentence embeddings using the sentence cosine similarity, on the
STS 2014 \cite{agirre2014semeval} and SICK 2014 \cite{marelli2014sick} datasets.
These similarity scores are compared to the gold-standard human judgements using Pearson's $r$ \cite{pearson1895note}
and Spearman's $\rho$ \cite{spearman1904proof} correlation scores.
The SICK dataset consists of about 10,000 sentence pairs along with relatedness scores of the pairs.
The STS 2014 dataset contains 3,770 pairs, divided into six different categories on the basis of the origin of sentences/phrases, namely Twitter, headlines, news, forum,
WordNet and images.

\section{Results and Discussion}
\label{sec:results}

In Tables \ref{sup-eval} and \ref{unsup-eval}, we compare our results with those obtained by \cite{Hill:2016uu} on different models.
Table \ref{macro_avg} in the last column shows the dramatic improvement in training time of our models (and other C-BOW-inspired models) in contrast to neural network based models.
All our Sent2Vec models are trained on a machine with 2x Intel Xeon E5$-$2680v3, 12 cores @2.5GHz.
\begin{table*}[h]
\centering
\captionsetup{font=small}
\small
\label{table1}
\begin{tabular}{|c|l|llllll|l|}
\hline
\multicolumn{1}{|l|}{Data}                                                                                                                           & Model                         & \begin{tabular}[c]{@{}c@{}}MSRP \\ (Acc / F1) \end{tabular}  				& MR   				& CR   				& SUBJ 				& MPQA 				& TREC 				& Average \\ \hline
\multirow{9}{*}{\begin{tabular}[c]{@{}c@{}}Unordered Sentences:\\ (Toronto Books; \\ 70 million sentences, \\ 0.9 Billion Words)\end{tabular}}       & SAE                           & \textbf{74.3} / {81.7}     		& 62.6 				& 68.0 				& 86.1 				& 76.8 				& 80.2 				& 74.7 \\
                                                                                                                                                     & SAE + embs.                   & 70.6 / 77.9     				& 73.2 				& 75.3 				& 89.8 				& 86.2 				& 80.4 				& 79.3\\
                                                                                                                                                     & SDAE                          & \underline{\textbf{76.4 / 83.4}}     	& 67.6 				& 74.0 				& 89.3 				& 81.3 				& 77.7 				& 78.3 \\
                                                                                                                                                     & SDAE + embs.                  & \textbf{73.7} / 80.7     		& 74.6 				& 78.0 				& 90.8 				& \textbf{86.9}			& 78.4 				& 80.4\\
                                                                                                                                                     & ParagraphVec DBOW             & 72.9 / 81.1     				& 60.2 				& 66.9 				& 76.3 				& 70.7 				& 59.4 				& 67.7\\
                                                                                                                                                     & ParagraphVec DM               & 73.6 / \textbf{81.9}     		& 61.5 				& 68.6 				& 76.4 				& 78.1 				& 55.8 				& 69.0 \\
                                                                                                                                                     & Skipgram                      & 69.3 / 77.2     				& 73.6 				& 77.3 				& 89.2 				& 85.0 				& 82.2 				& 78.5\\
                                                                                                                                                     & C-BOW                         & 67.6 / 76.1     				& 73.6 				& 77.3 				& 89.1 				& 85.0 				& 82.2 				& 79.1 \\
                                                                                                                                                     & Unigram TFIDF                 & 73.6 / 81.7     				& 73.7 				& 79.2 				& 90.3 				& 82.4 				& \textbf{85.0} 		& 80.7 \\
                                                                                                                                                     & \textbf{Sent2Vec uni.}        & 72.2 / 80.3     				& 75.1 				& {\textbf{80.2}} 		& 90.6 				& \textbf{86.3}  		& 83.8 				& \textbf{81.4} \\
                                                                                                                                                     & \textbf{Sent2Vec uni. + bi.}  & 72.5 / 80.8     				& \textbf{75.8} 		& \underline{\textbf{80.3}} 	& \textbf{91.2} 		& 85.9 				& \textbf{86.4} 		& \textbf{82.0} \\\hline
\multirow{3}{*}{\begin{tabular}[c]{@{}c@{}}Ordered Sentences:\\ Toronto Books\end{tabular}}                                                          & SkipThought                   & 73.0 / \textbf{82.0}     		& \underline{\textbf{76.5}}	& \textbf{80.1} 		& \underline{\textbf{93.6}} 	& \underline{\textbf{87.1}} 	& \underline{\textbf{92.2}} 	& \underline{\textbf{83.8}} \\
                                                                                                                                                     & FastSent                      & 72.2 / 80.3     				& 70.8 				& 78.4 				& 88.7 				& 80.6 				& 76.8 				& 77.9\\
                                                                                                                                                     & FastSent+AE                   & 71.2 / 79.1     				& 71.8 				& 76.7 				& 88.8 				& 81.5 				& 80.4 				& 78.4 \\ \hline
\multicolumn{1}{|l|}{2.8 Billion words}                                                                                                              & C-PHRASE                      & 72.2 / 79.6     				& \textbf{75.7} 		& 78.8 				& \textbf{91.1} 		& 86.2	 			& 78.8 				& 80.5 \\ \hline
\end{tabular}
\vspace{1mm}
\caption{Comparison of the performance of different models on different \textbf{supervised evaluation} tasks. An underline
indicates the best performance for the dataset. Top 3 performances in each data category are shown in bold. The average is calculated as the
average of accuracy for each category (For MSRP, we take the accuracy).
)}\label{sup-eval}
\end{table*}
Along with the models discussed in Section \ref{sec:rel-work}, this also includes the sentence embedding baselines obtained by simple averaging of word embeddings over the sentence, in both the C-BOW and skip-gram variants.
TF-IDF BOW is a representation consisting of the counts of the 200,000 most common feature-words, weighed by their TF-IDF frequencies.
To ensure coherence, we only include unsupervised models in the main paper. Performance of supervised and semi-supervised models on these evaluations can be observed in Tables \ref{sup-evalAppendix} and \ref{unsup-evalAppendix} in the supplementary material.

\textbf{Downstream Supervised Evaluation Results.}
On running supervised evaluations and observing the results in Table \ref{sup-eval}, we find that  on an average our models are second only to SkipThought vectors.
Also, both our models achieve state of the art results on the CR task.
We also observe that on half of the supervised tasks, our unigrams + bigram model is the best model after SkipThought.
Our models are weaker on the MSRP task (which consists of the identification of labelled paraphrases)
compared to state-of-the-art methods. However, we observe that the models which perform very strongly on this task end up faring very poorly on the other tasks, indicating a lack of generalizability.  %

On rest of the tasks, our models perform extremely well.  The SkipThought model is able to outperform our models on most of the tasks as it is trained to predict the previous and next
sentences and a lot of tasks are able to make use of this contextual information missing in our Sent2Vec models.
For example, the TREC task is a poor measure of how one predicts the content of the sentence (the question) but a good measure of
how  the next sentence in the sequence (the answer) is predicted.

\textbf{Unsupervised Similarity Evaluation Results.}
In Table \ref{unsup-eval}, we see that our Sent2Vec models are state-of-the-art on the majority of tasks when comparing to all the unsupervised models trained on the Toronto corpus, and clearly achieve the best averaged performance.
Our Sent2Vec models also on average outperform or are at par with the C-PHRASE model, despite significantly lagging behind on the STS 2014 WordNet and News subtasks.
This observation can be attributed to the fact that a big chunk of the data that the C-PHRASE model is trained on comes from
English Wikipedia, helping it to perform well on datasets involving definition and news items. Also, C-PHRASE uses data three times the size of the Toronto book corpus.
Interestingly, our model outperforms C-PHRASE when trained on Wikipedia, as shown in Table \ref{macro_avg}, despite the fact that we use no parse tree information.%

\textbf{Official STS 2017 benchmark.}
In the official results of the most recent edition of the STS 2017 benchmark \cite{Cer:2017wl}, our model also significantly outperforms C-PHRASE, and in fact delivers the best unsupervised baseline method.

\begin{table*}[]
\centering
\captionsetup{font=small}
\small
\label{table2}
\begin{tabular}{|l|llllll|l|l|}
\hline
                      		& \multicolumn{6}{c|}{STS 2014}                               																		& SICK 2014    & \\
Model                 		& News    			& Forum   				& WordNet 			& Twitter 			& Images  			& Headlines 			& Test + Train & Average\\ \hline
SAE                   		& .17/.16 			& .12/.12 				& .30/.23 			& .28/.22 			& .49/.46 			& .13/.11   			& .32/.31 			& .26/.23      \\
SAE + embs.           		& .52/.54 			& .22/.23 				& .60/.55 			& .60/.60 			& .64/.64 			& .41/.41   			& .47/.49 			& .50/.49     \\
SDAE                  		& .07/.04 			& .11/.13 				& .33/.24 			& .44/.42 			& .44/.38 			& .36/.36   			& .46/.46 			& .31/.29      \\
SDAE + embs.          		& .51/.54 			& .29/.29 				& .56/.50 			& .57/.58 			& .59/.59 			& .43/.44   			& .46/.46 			& .49/.49      \\
ParagraphVec DBOW     		& .31/.34 			& .32/.32 				& .53/.50 			& .43/.46 			& .46/.44 			& .39/.41   			& .42/.46 			& .41/.42      \\
ParagraphVec DM       		& .42/.46 			& .33/.34 				& .51/.48 			& .54/.57 			& .32/.30 			& .46/.47   			& .44/.40 			& .43/.43      \\
Skipgram              		& .56/.59 			& .42/.42 				& .73/\textbf{.70} 		& \underline{\textbf{.71}}/.74 	& .65/.67 			& .55/.58  			& .60/.69 			& .60/.63     \\
C-BOW                 		& .57/.61 			& \textbf{.43/.44} 			& .72/.69 			& \textbf{\underline{.71/.75}} 	& .71/.73 			& .55/.59  			& .60/.69 			& .60/.65       \\
Unigram TF-IDF        		& .48/.48 			& .40/.38 				& .60/.59 			& .63/.65 			& .72/.74 			& .49/.49 			& .52/.58 			& .55/.56       \\
\textbf{Sent2Vec uni.}		& \textbf{.62/.67} 		& \textbf{.49/.49} 			& \textbf{.75/.72} 		& .70/\textbf{\underline{.75}} 	& \textbf{\underline{.78/.82}} 	& \textbf{{.61}/.63} 		& \textbf{{.61}/.70} 		& \textbf{\underline{.65/.68}} \\
\textbf{Sent2Vec uni. + bi.}   	& \textbf{.62/.67} 		& \textbf{\underline{.51/.51}} 		& .71/.68 			& .70/\textbf{\underline{.75}} 	& \textbf{.75/.79} 		& {.59/.62} 			& \textbf{\underline{.62}/.70}  & \textbf{\underline{.65}/.67}  \\ \hline
SkipThought           		& .44/.45 			& .14/.15 				& .39/.34 			& .42/.43 			& .55/.60 			& .43/.44   			& .57/.60 			& .42/.43       \\
FastSent              		& .58/.59 			& .41/.36 				& \textbf{.74/.70} 		& .63/.66 			& .74/.78 			& .57/.59   			& \textbf{.61/\underline{.72}} 	& .61/.63      \\
FastSent+AE           		& .56/.59 			& .41/.40 				& .69/.64 			& .70/.74 			& .63/.65 			& .58/.60  			& .60/.65 			& .60/.61 \\
Siamese C-BOW\footnotemark      & .58/.59 			& .42/.41 				& .66/.61 			& \underline{\textbf{.71}}/.73  & .65/.65 			& \textbf{\underline{.63}/.64}  & $-$  				& $-$ \\ \hline
C-PHRASE               		& \textbf{\underline{.69/.71}} 	& \textbf{.43}/.41			& \textbf{\underline{.76/.73}} 	& .60/.65 			& \textbf{.75/.79} 		& \textbf{.60/\underline{.65}}  & .60/\textbf{\underline{.72}} 	& \textbf{.63/.67}      \\ \hline

\end{tabular}
\caption{\textbf{Unsupervised Evaluation Tasks}: Comparison of the performance of different models on Spearman/Pearson correlation measures. An underline
indicates the best performance for the dataset. Top 3 performances in each data category are shown in bold.  The average is calculated as the
average of entries for each correlation measure.%
 }\label{unsup-eval}
\end{table*}

\footnotetext{For the Siamese C-BOW model trained on the Toronto corpus, supervised evaluation as well as similarity evaluation results on the SICK 2014 dataset are unavailable.}

\textbf{Macro Average.}
To summarize our contributions on both supervised and unsupervised tasks, in Table \ref{macro_avg} we present the results in terms of the macro average over the averages of both supervised and unsupervised
tasks along with the training times of the models\footnote{time taken to train C-PHRASE models is unavailable}.
For unsupervised tasks, averages are taken over both Spearman and Pearson scores. The comparison includes the best performing unsupervised and semi-supervised methods described in Section \ref{sec:rel-work}.
For models trained on the Toronto books dataset, we report a 3.8 $\%$ points improvement over the state of the art.
Considering all supervised, semi-supervised methods and all datasets compared in \cite{Hill:2016uu}, we report a 2.2 $\%$ points improvement.

We also see a noticeable improvement in accuracy as we use larger datasets like Twitter and Wikipedia. We furthermore see that the Sent2Vec models are faster to train when compared to methods like SkipThought and DictRep, owing to the SGD optimizer allowing a high degree of parallelizability.

\begin{table*}[!htb]
\scriptsize
\captionsetup{font=small}
\centering
\begin{tabular}{|c|c|c|c|c|c|c|c|}\hline
\multicolumn{1}{|c|}{\begin{tabular}[c]{@{}c@{}}Type\end{tabular}} & \multicolumn{1}{c|}{\begin{tabular}[c]{@{}c@{}}Training corpus\end{tabular}} & Method & \multicolumn{1}{c|}{\begin{tabular}[c]{@{}c@{}}Supervised\\average\end{tabular}} & \multicolumn{1}{c|}{\begin{tabular}[c]{@{}c@{}}Unsupervised\\average\end{tabular}} & \multicolumn{1}{c|}{\begin{tabular}[c]{@{}c@{}}Macro\\average\end{tabular}} &  \multicolumn{1}{c|}{\begin{tabular}[c]{@{}c@{}}Training time \\  (in hours)\end{tabular}}\\ \hline
unsupervised & twitter (19.7B words) &  \begin{tabular}[c]{@{}c@{}}\textbf{Sent2Vec uni. + bi.}\end{tabular}  & 83.5  & 68.3  & 75.9 & 6.5*  \\ \hline %
unsupervised & twitter (19.7B words) & \begin{tabular}[c]{@{}c@{}}\textbf{Sent2Vec uni.} \end{tabular}  & 82.2  & 69.0  & 75.6 & 3* \\ \hline %
unsupervised & Wikipedia (1.7B words) & \begin{tabular}[c]{@{}c@{}}\textbf{Sent2Vec uni. + bi.}\end{tabular}  & 83.3  & 66.2  & 74.8 & 2* \\ \hline %
unsupervised & Wikipedia (1.7B words) & \begin{tabular}[c]{@{}c@{}}\textbf{Sent2Vec uni.} \end{tabular}  & 82.4  & 66.3  & 74.3 & 3.5* \\  \hline %
unsupervised & Toronto books (0.9B words) & \begin{tabular}[c]{@{}c@{}}\textbf{Sent2Vec books uni.} \end{tabular}  & 81.4  & 66.7  & 74.0 & 1*  \\ \hline %
unsupervised & Toronto books (0.9B words) &\begin{tabular}[c]{@{}c@{}}\textbf{Sent2Vec books uni. + bi.}\end{tabular}  & 82.0  & 65.9  & 74.0 & 1.2* \\ \hline %
semi-supervised & structured dictionary dataset & \begin{tabular}[c]{@{}c@{}}DictRep BOW + emb\end{tabular}  & 80.5  & 66.9  & 73.7 & 24** \\ \hline
unsupervised & 2.8B words + parse info. & \begin{tabular}[c]{@{}c@{}}C-PHRASE\end{tabular}  & 80.5  & 64.9  & 72.7 &  $-$ \\  \hline
unsupervised & Toronto books (0.9B words)  & \begin{tabular}[c]{@{}c@{}}C-BOW\end{tabular}  & 79.1  & 62.8  & 70.2 &   2 \\ \hline
unsupervised & Toronto books (0.9B words)  &\begin{tabular}[c]{@{}c@{}}FastSent\end{tabular}  & 77.9  & 62.0  & 70.0 &  2 \\ \hline
unsupervised & Toronto books (0.9B words)  &\begin{tabular}[c]{@{}c@{}}SkipThought\end{tabular}  & 83.8  & 42.5  & 63.1 & 336** \\ \hline
\end{tabular}
\caption{Best unsupervised and semi-supervised methods ranked by macro average along with their training times.
**~indicates trained on GPU. * indicates trained on a single node using 30 threads.
Training times for non-Sent2Vec models are due to \citet{Hill:2016uu}. For CPU based competing methods, we were able to reproduce all published timings (+-10\%) using our same hardware as for training Sent2Vec.
 }\label{macro_avg}
\end{table*}

We can clearly see Sent2Vec outperforming other unsupervised and even semi-supervised methods.
This can be attributed to the superior generalizability of our model across supervised and unsupervised tasks.

\begin{table*}[!htb]
\centering
\captionsetup{font=small}
\small
\begin{tabular}{|l|c|c|c|c|}
\hline
Dataset              & \multicolumn{1}{c|}{\begin{tabular}[c]{@{}c@{}}Unsupervised \\ GloVe ({840B words}) \\ + WR\end{tabular}} & \multicolumn{1}{c|}{\begin{tabular}[c]{@{}c@{}}Semi-supervised \\ PSL + WR\end{tabular}} & \multicolumn{1}{c|}{\begin{tabular}[c]{@{}c@{}}Sent2Vec Unigrams \\({19.7B words})\\ Tweets Model\end{tabular}} & \multicolumn{1}{c|}{\begin{tabular}[c]{@{}c@{}}Sent2Vec Unigrams + Bigrams \\({19.7B words})\\ Tweets Model\end{tabular}} \\ \hline
STS 2014             & 0.685                                                                                                     & 0.735                                                                                    & 0.710                                                                                                           & 0.701                                                                                                                     \\
SICK 2014            & 0.722                                                                                                     & 0.729                                                                                    & 0.710                                                                                                           & 0.715                                                                                                                     \\ \hline
Supervised average   & 0.815                                                                                                     & 0.807                                                                                    & 0.822                                                                                                           & 0.835                                                                                                                     \\ \hline
\end{tabular}
\caption{Comparison of the performance of the unsupervised and semi-supervised sentence embeddings by \cite{arora2017}
 with our models. Unsupervised comparisons are in terms of Pearson's correlation, while comparisons
 on supervised tasks are stating the average described in Table \ref{sup-eval}.}
 \label{arora-table}
\end{table*}

\textbf{Comparison with \citet{arora2017}.}
We also compare our work with \citet{arora2017} who also use additive compositionality to obtain sentence embeddings.
However, in contrast to our model, they use fixed, pre-trained word embeddings to build a weighted
average of these embeddings using unigram probabilities. While we couldn't find pre-trained state of the
art word embeddings trained on the Toronto books corpus, we evaluated
their method using GloVe embeddings obtained from the larger Common Crawl Corpus\footnote{\url{http://www.cs.toronto.edu/~mbweb/}},
which is 42 times larger than our twitter corpus, greatly favoring their method over ours.

In Table \ref{arora-table}, we report an experimental comparison to their model on unsupervised tasks. %
In the table, the suffix W indicates that their down-weighting scheme has been used, while the suffix~R indicates
the removal of the first principal component.
They report values of $a \in [10^{-4},10^{-3}]$ as giving the best results and
used $a=10^{-3}$ for all their experiments. We observe that our results are
competitive with the embeddings of \citet{arora2017} for purely unsupervised methods.
It is important to note that the scores obtained from supervised task-specific PSL embeddings trained for the purpose of
semantic similarity outperform our method on both SICK and average STS 2014,
which is expected as our model is trained purely unsupervised.

In order to facilitate a more detailed comparison, we also evaluated the unsupervised Glove + WR embeddings on downstream
 supervised tasks and compared them to our twitter models. To use \citet{arora2017}'s method in a supervised setup, we precomputed
 and stored the common discourse vector $\cv_0$ using 2 million random Wikipedia sentences.
On an average, our models outperform their unsupervised models by a significant margin, this despite the fact that they
used GloVe embeddings trained on larger corpora than ours (42 times larger).
Our models also outperform their semi-supervised PSL + WR model.
This indicates our model learns a more precise weighing scheme than the static one proposed by \citet{arora2017}.

\vspace{0.1mm}
\begin{figure}[!htb]
\hspace*{-1.0cm}
\captionsetup{font=small}
  \includegraphics[width=1.20\columnwidth]{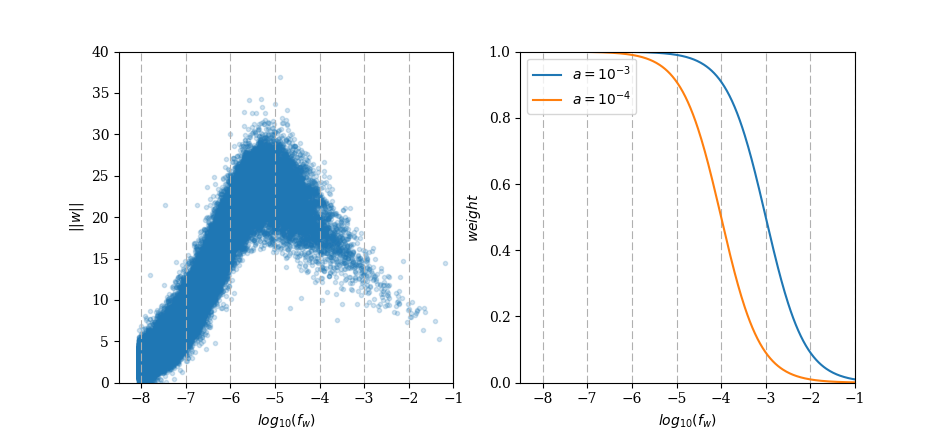}
\vspace{-6mm}
\caption{\emph{Left figure:} the profile of the word vector $L_2$-norms as a function of $\log(f_w)$ for each vocabulary word $w$, as learnt by our unigram model trained on Toronto books.
\emph{Right figure:} down-weighting scheme proposed by \citet{arora2017}: $weight(w) = \frac{a}{a+f_w}$.}
\label{Arora_comparison}
\end{figure}
\textbf{The effect of datasets and n-grams.} 
Despite being trained on three very different datasets, all of our models generalize well to sometimes very specific domains.
Models trained on Toronto Corpus are the state-of-the-art on the STS 2014 images dataset even beating the supervised CaptionRep model %
trained on images.
We also see that addition of bigrams to our models doesn't help much
when it comes to unsupervised evaluations but gives a significant boost-up in accuracy on supervised tasks.
We attribute this phenomenon to the ability of bigrams models to capture some non-compositional features missed by unigrams models.
Having a single representation for \textit{``not good"} or \textit{``very bad"} can boost the supervised model's ability to infer relevant features for the corresponding classifier.
For semantic similarity tasks however, the relative uniqueness of bigrams results in pushing sentence representations further apart, which can explain the average drop of scores for bigrams models on those tasks.

\textbf{On learning the importance and the direction of the word vectors}. Our model -- by learning how to generate and compose word
vectors -- has to learn both the direction of the word embeddings as well as  their norm. Considering the norms of the used word vectors as by our averaging over the sentence, we observe an interesting distribution of the ``importance'' of each word.
In Figure \ref{Arora_comparison} we show the profile of the $L_2$-norm as a function of $\log(f_w)$ for each $w\in \vocab$, and compare it to the static down-weighting mechanism of \citet{arora2017}.
We can observe that our model is learning to down-weight frequent tokens by itself. It is also
down-weighting rare tokens and the $\mathop{norm}$ profile seems to roughly follow Luhn's hypothesis \cite{luhn1958automatic},
a well known information retrieval paradigm, stating that mid-rank terms are the most significant to discriminate content.

\vspace{-1.3mm}
\section{Conclusion}
\vspace{-1.3mm}

In this paper, we introduce a novel, computationally efficient, unsupervised, C-BOW-inspired method to train and infer sentence embeddings.
On supervised evaluations, our method, on an average, achieves better performance than all other unsupervised competitors with the exception of SkipThought. However, SkipThought vectors show a very poor performance on sentence similarity tasks while our model is state-of-the-art for these evaluations on average.
Also, our model is generalizable, extremely fast to train, simple to understand and easily interpretable, showing the relevance of simple and well-grounded representation models in contrast to the models using deep architectures.  Future work could focus on augmenting the model to
exploit data with ordered sentences. Furthermore,
we would like to investigate the model's ability to use pre-trained
embeddings for downstream transfer learning tasks.

\subsubsection*{Acknowledgments}
We are indebted to Piotr Bojanowski and Armand Joulin for helpful discussions.
This project was supported by a Google Faculty Research Award.

\bibliography{unsup-fasttext,bib-mj}
\bibliographystyle{acl_natbib}
\newpage
\appendix

\clearpage
\onecolumn
\begin{center}
{\centering \LARGE Supplementary Material}
\end{center}

\section{Parameters for training models}
\begin{table*}[!htb]
\scriptsize

\centering
\begin{tabular}{|c|l|l|l|l|l|l|l|l|}
\hline
Model                                                                          & \multicolumn{1}{c|}{\begin{tabular}[c]{@{}c@{}}Embedding\\ Dimensions\end{tabular}} & \multicolumn{1}{c|}{\begin{tabular}[c]{@{}c@{}}Minimum\\ word count\end{tabular}} & \multicolumn{1}{c|}{\begin{tabular}[c]{@{}c@{}}Minimum\\ Target word \\ Count\end{tabular}} & \multicolumn{1}{c|}{\begin{tabular}[c]{@{}c@{}}Initial\\ Lear ning\\ Rate\end{tabular}} & \multicolumn{1}{c|}{\begin{tabular}[c]{@{}c@{}}Epochs\end{tabular}} & \multicolumn{1}{c|}{\begin{tabular}[c]{@{}c@{}}Subsampling\\ hyper-parameter\end{tabular}} & \multicolumn{1}{c|}{\begin{tabular}[c]{@{}c@{}}Bigrams\\ Dropped\\ per sentence\end{tabular}} & \multicolumn{1}{c|}{\begin{tabular}[c]{@{}c@{}}Number of\\ negatives\\ sampled\end{tabular}} \\ \hline
\begin{tabular}[c]{@{}c@{}}Book corpus \\ Sent2Vec\\ unigrams\end{tabular}               & 700                                                                                 & 5                                                                                 & 8                                                                                          & 0.2                                                                          & 13                                                                                               & $1 \times 10^{-5}$                                                                 & -                                                                                             & 10                                                                                           \\ \hline
\begin{tabular}[c]{@{}c@{}}Book corpus \\ Sent2Vec \\ unigrams + bigrams\end{tabular}    & 700                                                                                 & 5                                                                                 & 5                                                                                          & 0.2                                                                          & 12                                                                                               & $5 \times 10^{-6}$                                                               & 7                                                                                             & 10                                                                                           \\ \hline
\begin{tabular}[c]{@{}c@{}}Wiki Sent2Vec\\ unigrams\end{tabular}               & 600                                                                                 & 8                                                                                 & 20                                                                                          & 0.2                                                                          & 9                                                                                               & $1 \times 10^{-5}$                                                                 & -                                                                                             & 10                                                                                           \\ \hline
\begin{tabular}[c]{@{}c@{}}Wiki Sent2Vec \\ unigrams + bigrams\end{tabular}    & 700                                                                                 & 8                                                                                 & 20                                                                                          & 0.2                                                                          & 9                                                                                               & $5 \times 10^{-6}$                                                               & 4                                                                                             & 10                                                                                           \\ \hline
\begin{tabular}[c]{@{}c@{}}Twitter Sent2Vec\\  unigrams\end{tabular}           & 700                                                                                 & 20                                                                                & 20                                                                                          & 0.2                                                                          & 3                                                                                               & $1 \times 10^{-6}$                                                                 & -                                                                                             & 10                                                                                           \\ \hline
\begin{tabular}[c]{@{}c@{}}Twitter Sent2Vec\\  unigrams + bigrams\end{tabular} & 700                                                                                 & 20                                                                                & 20                                                                                          & 0.2                                                                          & 3                                                                                               & $1 \times 10^{-6}$                                                                 & 3                                                                                             & 10                                                                                           \\ \hline
\end{tabular}
\caption{Training parameters for the Sent2Vec models}\label{table_train}
\end{table*}

\section{L1 regularization of models}\label{l1Details}
Optionally, our model can be additionally improved by adding an L1 regularizer term in the objective function, leading to slightly better generalization performance. Additionally, encouraging sparsity in the embedding vectors is  beneficial for memory reasons, allowing higher embedding dimensions $h$.

We propose to apply L1 regularization individually to each word (and n-gram) vector (both source and target vectors).
Formally, the training objective function \eqref{eq:trainObj} then
becomes
\begin{align}\label{eq:trainObjL1}
 \min_{\Um,\Vm} \sum_{S \in \corp} \sum_{w_t \in S} &
q_p(w_t) \bigg( \Big(\ell\big(\uv_{w_t}^\top \vv_{S\setminus\{w_t\}}\big) + \tau(\|\uv_{w_t}  \|_1 + \|\vv_{S\setminus\{w_t\}}  \|_1)\Big) + \\
&\!\!\!  |N_{w_t}| \sum_{w' \in \vocab} \ \
q_n(w') \Big(\ell\big(\!-\uv_{w'}^\top \vv_{S\setminus\{w_t\}}\big) + \tau(\|\uv_{w'}  \|_1) \Big) \bigg) \notag
\end{align}
where $\tau$ is the regularization parameter.

Now, in order to minimize a function of the form $f(\textbf{z}) + g(\textbf{z})$ where $g(\textbf{z})$ is not differentiable over the domain, we can use the basic proximal-gradient
scheme. In this iterative method, after doing a gradient descent step on $f(\textbf{z})$ with learning rate $\alpha$, we update $\textbf{z}$ as
\begin{align}
 \textbf{z}_{n+1} = prox_{\alpha,g}(\textbf{z}_{n + \frac{1}{2}})
\end{align}

where $prox_{\alpha,g}(\textbf{x}) = \argmin_\textbf{y} \{ g(\textbf{y}) +  \frac{1}{2\alpha} \|\textbf{y} -\textbf{x}\|_2^2\}$ is called the proximal function \cite{rockafellar1976monotone} of $g$
with $\alpha$ being the proximal parameter and $\textbf{z}_{n + \frac{1}{2}}$ is the value of $\textbf{z}$ after a gradient (or SGD)
step on $\textbf{z}_n$.

In our case, $g(\textbf{z}) = \|\textbf{z}\|_1$ and the corresponding proximal operator is given by
\begin{align}
prox_{\alpha,g}(\textbf{x}) = sign(\textbf{x}) \odot \max(|\textbf{x}_{n}| - {\alpha},0)
\end{align}
where $\odot$ corresponds to element-wise product.

Similar to the proximal-gradient scheme, in our case we can optionally use the thresholding operator on the updated word
and n-gram vectors after an SGD step. The soft thresholding parameter used for this update is $\frac{\tau \cdot lr'}{|R(S\setminus\{w_t\})| }$ and ${\tau \cdot lr'}$
for the source and target vectors respectively where $lr'$ is the
current learning rate, $\tau$ is the $L1$ regularization parameter and $S$ is the sentence on which SGD is being run. \\

We observe that $L1$ regularization
using the proximal step gives our models a small boost in performance. Also, applying the thresholding operator takes only $|R(S\setminus\{w_t\})| \cdot h$ floating point
operations for the updating the word vectors corresponding to the sentence and $(|N| + 1) \cdot h$ for updating the target as well as the negative word vectors, where
$|N|$ is the number of negatives sampled and $h$ is the embedding dimension.
Thus, performing $L1$ regularization using soft-thresholding operator comes with a small computational overhead.

We set $\tau$ to be 0.0005 for both the Wikipedia and the Toronto Book Corpus unigrams + bigrams models.

\section{Performance comparison with Sent2Vec models trained on different corpora}
\begin{table*}[!htb]
\centering
\scriptsize
\label{table4}
\begin{tabular}{|c|l|llllll|c|}
\hline
\multicolumn{1}{|l|}{Data}                                                                                                                           & Model                         & MSRP (Acc / F1) 		& MR   			& CR  		 	& SUBJ 		& MPQA 			& TREC 			& Average\\ \hline
\multirow{4}{*}{\begin{tabular}[c]{@{}c@{}}Unordered Sentences:\\ (Toronto Books)\end{tabular}}       						     & \textbf{Sent2Vec uni.}        & 72.2 / 80.3     		& 75.1 			& \textbf{80.2} 	& 90.6 		& 86.3 			& 83.8        		& 81.4 \\
\multicolumn{1}{|l|}{}                                                                                                                               & \textbf{Sent2Vec uni. + bi.}  & 72.5 / 80.8     		& 75.8 			& \textbf{80.3}		& 91.2 		& 85.9 			& 86.4        		& 82.0 \\
\multicolumn{1}{|l|}{}                                                                                                                               & \textbf{Sent2Vec uni. + bi. $L1$-reg} & 71.6 / 80.1  	& 76.1 			& \textbf{80.9}		& 91.1 		& 86.1 			& 86.8 	 		& 82.1 \\\hline
\multicolumn{1}{|l|}{\multirow{4}{*}{\begin{tabular}[c]{@{}l@{}}Unordered sentences: Wikipedia \\(69 million sentences; 1.7 B words)\end{tabular}}}  & \textbf{Sent2Vec uni.}        & 71.8 / 80.2     		& \textbf{77.3}		& \textbf{80.3}		& 92.0 		& \textbf{\underline{87.4}} 	& 85.4 			& 82.4 \\
\multicolumn{1}{|l|}{}																     & \textbf{Sent2Vec uni. + bi.}  & 72.4 / 80.8     		& \textbf{77.9}		& \textbf{80.9}		& 92.6 		& 86.9 			& 89.2 			& 83.3 \\
\multicolumn{1}{|l|}{}																     & \textbf{Sent2Vec uni. + bi. $L1$-reg}  & 73.6 / 81.5 	& \textbf{\underline{78.1}} 	& \textbf{81.5}		& 92.8 		& \textbf{87.2}		& 87.4 			& 83.4 \\ \hline
\multicolumn{1}{|l|}{\multirow{2}{*}{\begin{tabular}[c]{@{}l@{}}Unordered sentences: Twitter \\(1.2 billion sentences; 19.7 B words)\end{tabular}}}  & \textbf{Sent2Vec uni.}        & 71.5 / 80.0     		& \textbf{77.1}		& \textbf{81.3}		& 90.8 		& \textbf{87.3}		& 85.4 			& 82.2 \\
\multicolumn{1}{|l|}{}                                                                                                                               & \textbf{Sent2Vec uni. + bi.}  & 72.4 / 80.6     		& \textbf{78.0}		& \textbf{\underline{82.1}} 	& 91.8 		& 86.7 			& 89.8 			& 83.5 \\ \hline
\multirow{6}{*}{\begin{tabular}[c]{@{}c@{}}Other structured\\  Data Sources\end{tabular}}                                                            & CaptionRep BOW                & 73.6 / 81.9     		& 61.9 			& 69.3 			& 77.4 		& 70.8 			& 72.2 			& 70.9 \\
                                                                                                                                                     & CaptionRep RNN                & 72.6 / 81.1     		& 55.0 			& 64.9 			& 64.9 		& 71.0 			& 62.4 			& 65.1 \\
                                                                                                                                                     & DictRep BOW                   & 73.7 / 81.6     		& 71.3 			& 75.6 			& 86.6 		& 82.5 			& 73.8 			& 77.3 \\
                                                                                                                                                     & DictRep BOW+embs              & 68.4 / 76.8     		& 76.7 			& 78.7 			& 90.7 		& 87.2 			& 81.0 			& 80.5 \\
                                                                                                                                                     & DictRep RNN                   & 73.2 / 81.6     		& 67.8 			& 72.7 			& 81.4 		& 82.5 			& 75.8 			& 75.6 \\
                                                                                                                                                     & DictRep RNN+embs.             & 66.8 / 76.0     		& 72.5 			& 73.5 			& 85.6 		& 85.7 			& 72.0 			& 76.0 \\ \hline

\end{tabular}

\normalsize
\caption{Comparison of the performance of different Sent2Vec models with different semi-supervised/supervised models
on different \textbf{downstream supervised evaluation} tasks. An underline
indicates the best performance for the dataset and Sent2Vec model performances are bold if they perform as well or better than
all other non-Sent2Vec models, including those presented in Table \ref{sup-eval}.}\label{sup-evalAppendix}
\end{table*}

\begin{table*}[!htb]
\centering
\scriptsize
\label{table5}
\begin{tabular}{|l|llllll|l|l|}
\hline
                      & \multicolumn{6}{c|}{STS 2014}                                         & SICK 2014 																					& Average   \\
Model                 				& News    			& Forum   			& WordNet 			& Twitter 			& Images  			& Headlines 			&  Test + Train & \\ \hline
{Sent2Vec book corpus uni.}			& .62/.67 			& \textbf{.49/.49}		& .75/.72. 			& .70/\textbf{.75}		& \textbf{\underline{.78/.82}}	& .61/.63 			&  \textbf{.61}/.70 		& \textbf{.65}/.68 \\
{Sent2Vec book corpus uni. + bi.}   		& .62/.67 			& \textbf{.51/.51}		& .71/.68 			& .70/\textbf{.75}		& .75/.79 			& .59/.62 			& \textbf{.62}/.70 		& \textbf{.65}/.67 \\
{Sent2Vec book corpus uni. + bi. $L1$-reg}  	& .62/.68 			& \textbf{.51/.52}		& .72/.70			& .69/\textbf{.75}		& .76/.81	                & .60/.63 			& \textbf{.62}/.71		& \textbf{.66}/.68 \\ \hline
{Sent2Vec wiki uni.}              		& .66/.71 			& .47/\textbf{.47}		& .70/.68 			& .68/.72 			& .76/.79 			& \textbf{.63/.67} 		& \textbf{\underline{.64}}/.71	& \textbf{.65}/.68       \\
{Sent2Vec wiki uni. + bi.}    			& .68/\textbf{.74}		& \textbf{.50/.50}		& .66/.64 			& .67/.72 			& .75/.79 			& .62/\textbf{.67}		& \textbf{.63}/.71 		& \textbf{.65}/.68  \\
{Sent2Vec wiki uni. + bi. $L1$-reg}    		& \underline{\textbf{.69/.75}}	& \textbf{.52/.52}		& .72/.69 			& .67/.72 			& .76/.80 			& .61/\textbf{.66}		& \textbf{.63/\underline{.72}}	& \textbf{.66}/.69  \\ \hline
{Sent2Vec twitter uni.}           		& .67/\textbf{.74}		& \textbf{.52/.53}		& .75/.72 			& \textbf{\underline{.72/.78}}	& .77/.81			& \textbf{\underline{.64/.68}}	& \textbf{.62}/.71 		& \textbf{\underline{.67/.71}}  \\
{Sent2Vec twitter uni. + bi. } 			& .68/\textbf{.74}		& \textbf{\underline{.54/.54}}	& .72/.69 			& .70/\textbf{.77}		& .76/.79 			& .62/\textbf{.67}		& \textbf{.63/\underline{.72}}	& \textbf{.66/.70}  \\ \hline
CaptionRep BOW        				& .26/.26 			& .29/.22 			& .50/.35 			& .37/.31 			& \underline{.78}/.81		& .39/.36 			& .45/.44 			& .54/.62        \\
CaptionRep RNN        				& .05/.05 			& .13/.09 			& .40/.33 			& .36/.30 			& .76/\underline{.82}		& .30/.28 			& .36/.35  			& .51/.59       \\
DictRep BOW           				& .62/.67 			& .42/.40 			& .81/.81			& .62/.66			& .66/.68 			& .53/.58 			& .61/.63  			& .58/.66      \\
DictRep BOW + embs.    				& .65/.72 			& .49/.47 			& \underline{.85/.86}		& .67/.72 			& .71/.74 			& .57/.61 			& .61/.70  			& .62/.70       \\
DictRep RNN           				& .40/.46 			& .26/.23 			& .78/.78 			& .42/.42 			& .56/.56 			& .38/.40  			& .47/.49  			& .49/.55        \\
DictRep RNN + embs.   				& .51/.60 			& .29/.27 			& .80/.81 			& .44/.47 			& .65/.70 			& .42/.46  			& .52/.56  			& .49/.59        \\ \hline

\end{tabular}
\normalsize
\caption{  \textbf{Unsupervised Evaluation}: Comparison of the performance of different Sent2Vec models with
semi-supervised/supervised models on Spearman/Pearson correlation measures. An underline
indicates the best performance for the dataset and Sent2Vec model performances are bold if they perform as well or better than
all other non-Sent2Vec models, including those presented in Table \ref{unsup-eval}.
}\label{unsup-evalAppendix}
\end{table*}
\section{Dataset Description}
\normalsize
\begin{table*}[!htb]
\centering
\scriptsize
\begin{tabular}{|l|l|l|l|l|l|l|l|l|l|l|}
\hline
                   & \multicolumn{6}{c|}{STS 2014}                          & SICK 2014    & \multicolumn{1}{c|}{\multirow{2}{*}{\begin{tabular}[c]{@{}c@{}}Wikipedia\\ Dataset\end{tabular}}} & \multicolumn{1}{c|}{\multirow{2}{*}{\begin{tabular}[c]{@{}c@{}}Twitter\\ Dataset\end{tabular}}} & \multicolumn{1}{c|}{\multirow{2}{*}{\begin{tabular}[c]{@{}c@{}}Book Corpus\\ Dataset\end{tabular}}} \\
Sentence Length    & News  & Forum & WordNet & Twitter & Images & Headlines & Test + Train & \multicolumn{1}{c|}{}                                                                             & \multicolumn{1}{c|}{}                                                                           & \multicolumn{1}{c|}{}  \\ \hline
Average            & 17.23 & 10.12 & 8.85    & 11.64   & 10.17  & 7.82      & 9.67         & 25.25                                                                                             & 16.31                                                                                           & 13.32 \\ \hline
Standard Deviation & 8.66  & 3.30  & 3.10    & 5.28    & 2.77   & 2.21      & 3.75         & 12.56                                                                                             & 7.22                                                                                            & 8.94  \\ \hline
\end{tabular}
\caption{Average sentence lengths for the datasets used in the comparison.}\label{sent-lengths}
\end{table*}

\end{document}